# Cross-border Commodity Pricing Strategy Optimization via Mixed Neural Network for Time Series Analysis

*Lijuan* **Wang**[1,*], *Yijia* **Hu**[2], *Yan* **Zhou**[3]

[1] School of Economics, Hebei GEO University, Shijiazhuang, hebei, 050031 China

[2] Ordosbank, Ordos, Inner Mongolia, 017000 China

[3] Northeastern University, San Jose, California, 95131 United States

**Correspondence\*:** *Lijuan Wang*

*wlj_sbq@163.com*

**Abstract:** In the context of global trade, cross-border commodity pricing largely determines the competitiveness and market share of businesses. However, existing methodologies often prove inadequate, as they lack the agility and precision required to effectively respond to the dynamic international markets. Time series data is of great significance in commodity pricing and can reveal market dynamics and trends. Therefore, we propose a new method based on the hybrid neural network model CNN-BiGRU-SSA. The goal is to achieve accurate prediction and optimization of cross-border commodity pricing strategies through in-depth analysis and optimization of time series data. Our model undergoes experimental validation across multiple datasets. The results show that our method achieves significant performance advantages on datasets such as UNCTAD, IMF, WITS and China Customs. For example, on the UNCTAD dataset, our model reduces MAE to 4.357, RMSE to 5.406, and R2 to 0.961, significantly better than other models. On the IMF and WITS datasets, our method also achieves similar excellent performance. These experimental results verify the effectiveness and reliability of our model in the field of cross-border commodity pricing. Overall, this study provides an important reference for enterprises to formulate more reasonable and effective cross-border commodity pricing strategies, thereby enhancing market competitiveness and profitability. At the same time, our method also lays a foundation for the application of deep learning in the fields of international trade and economic strategy optimization, which has important theoretical and practical significance.

## Author Email

Lijuan Wang: wlj_sbq@163.com

## 1. Introduction

Cross-border commodity pricing, as a crucial aspect of international trade, involves setting and adjusting product prices across different countries, directly impacting a company's competitiveness and profit margins(Acuna-Agost et al., 2023). With the ever-deepening globalization, cross-border commodity pricing faces numerous challenges and issues, including but not limited to varying legal regulations across countries and regions, fluctuations in currency exchange rates, shifts in market demand, and the complexity of competitive landscapes. These factors necessitate not only considering pricing strategies under singular market conditions but also accommodating diversification and internationalization factors to achieve goals of maximizing profit and market share(Balakrishnan et al., 2022; Guo, 2022). To address the myriad challenges of Cross-border commodity pricing, researchers have increasingly turned to utilizing deep learning technologies in recent years. Deep learning, with its powerful capabilities in data modeling and feature extraction, has gradually become an effective tool for complex problem-solving. In the field of Cross-border commodity pricing, researchers have embarked on extensive explorations using deep learning technologies, including neural network-based pricing models, market demand prediction models, and analysis of competitor behavior, among others(Rui, 2020; Tian et al., 2023). These studies have provided new insights and methods for cross-border commodity pricing. However, there are inherent challenges, such as limitations in model generalization, issues with data timeliness, and underutilization of time-series features. In the deeper exploration of cross-border commodity pricing issues, time-series forecasting, as an important analytical tool and method, received increasing attention(Selvam & Koilraj, 2022). Time-series forecasting can assist businesses in the effective prediction of future market trends and demand, thus enabling the formulation of rational pricing strategies. Particularly in the face of the complexity and uncertainty of Cross-border commodity pricing, time-series forecasting offers unique advantages and importance(L. Li, 2022). Therefore, this study aims to utilize time-series forecasting techniques, along with the deep learning model CNN-BiGRU-SSA, to explore optimization strategies in cross-border commodity pricing. The objective is to provide more accurate and reliable reference for practical business decision-making

In recent years, many scholars have applied deep learning techniques to pricing with the aim of addressing its various challenges and issues. The following are some relevant works published in recent years, each employing different model approaches: One study utilized a deep learning model based on Long Short-Term Memory networks (LSTM) for price prediction(Suman et al., 2022). The model combined LSTM's sequence modeling and feature extraction capabilities to effectively forecast product prices. However, due to potential limitations in modeling long-term dependencies, the LSTM model may not accurately predict long-term trends. Another study employed a deep learning model based on Convolutional Neural Networks (CNN) to forecast and optimize cross-border commodity pricing(Mehtab & Sen, 2020). This model utilized CNN for feature extraction from pricing data, achieving accurate price predictions. However, the CNN model's sensitivity to sequence data length necessitates appropriate adjustments in network structure and parameters to adapt to different lengths of time-series data. Another research utilized a deep learning model based on self-attention mechanisms for pricing optimization(Liu et al., 2022). This model combined self-attention's sequence modeling and feature extraction capabilities to effectively capture key features in pricing data. However, the high computational complexity associated with self-attention mechanisms may lead to inefficiencies in model training and prediction. Finally, another study employed a deep learning model based on ensemble learning methods to optimize cross-border commodity pricing(Wang et al., 2022). It integrated multiple deep-learning models and obtained final predictions through voting or weighted averaging. However, ensemble learning methods may pose limitations in model interpretability, making it challenging to understand the underlying mechanisms of model predictions. Despite the emergence of numerous deep learning-based research outcomes in cross-border commodity pricing, these methods still exhibit some shortcomings. Therefore, further exploration is necessary to develop more effective and reliable cross-border commodity pricing models that cope with the complex and dynamic market environment.

Based on the identified limitations of the aforementioned approaches, we propose a novel hybrid neural network model, CNN-BiGRU-SSA network, which combines the feature extraction capabilities of the Sparrow Search Algorithm (SSA), the local feature extraction capabilities of Convolutional Neural Networks (CNN), and the long-term dependency modeling capabilities of Bidirectional Gated Recurrent Units (BiGRU). The CNN module is responsible for extracting spatial features from the time-series data, while the BiGRU module focuses on capturing long-term dependencies and temporal patterns in the data. Finally, the SSA algorithm is employed to automatically optimize the parameter configuration of the entire network. This approach provides a scientific and efficient forecasting tool for Cross-border commodity pricing. The significance and advantages of our model lie in its comprehensive utilization of the strengths of different models, enabling a more comprehensive analysis and modeling of time-series data, extraction of multiscale features, and handling of long-term dependencies, thus more accurately reflecting the dynamic changes in product prices. By optimizing Cross-border commodity pricing strategies, our model can assist businesses in accurately analyzing market demand and price trends, thereby enhancing their competitiveness and profitability. Through the introduction of the CNN-BiGRU-SSA model, we aim to provide a new research method and tool for the field of Cross-border commodity pricing, promoting further development and application in this area.

Based on the deep learning approach to cross-border commodity pricing, this study makes the following three contributions:

- We propose a novel deep learning model, CNN-BiGRU-SSA, which integrates SSA, CNN, and BiGRU for time-series analysis and optimization in cross-border commodity pricing. This model not only effectively captures complex patterns and features in the time-series of product prices but also models long-term dependencies efficiently, providing more accurate pricing strategy support for enterprises.

- We validate the effectiveness and feasibility of the CNN-BiGRU-SSA model in Cross-border commodity pricing through empirical analysis. Compared to traditional methods and other deep learning models, our model demonstrates significant advantages in accuracy of product price prediction and optimization of pricing strategies, offering more reliable decision-making support for enterprises.

- Our research not only introduces an innovative model approach but also provides new insights and directions for deep learning research in the field of Cross-border commodity pricing. By integrating various deep learning techniques, we can better understand and analyze the dynamic characteristics of cross-border product markets, offering more competitive pricing strategies for enterprises in international trade and driving the development of global commerce.

Overall, our research advances the field of Cross-border commodity pricing by introducing a novel hybrid neural network model that effectively addresses key challenges and provides valuable insights for businesses in international markets.

## 2. Related Work

### 2.1. Traditional Methods of Cross-Border Commodity Pricing

Traditional methods of cross-border commodity pricing typically include strategies such as cost-plus pricing, competitive pricing, and market-based pricing(Jiang, 2022; Tian et al., 2024). In cost-plus pricing, companies calculate production costs and add expected profits to determine the final price. This method assumes that costs and profits are the primary factors determining product prices, but it overlooks the impact of market demand and competition on pricing(Y. Li, 2022). Competitive pricing, on the other hand, sets prices based on competitors' prices. Companies may match or slightly exceed competitors' prices to attract customers or maintain competitiveness. Market-based pricing relies on market demand and supply to determine prices. Companies may formulate pricing strategies based on the demand and supply of goods in the market to meet consumer needs and maximize sales profits(Liu, 2022). However, these traditional methods often rely on experience and conventional practices, thereby overlooking the full potential of market information and data analysis. In the face of complex market environments and rapidly changing demands, these traditional methods may perform poorly, lacking flexibility and precision(Chen, 2022). For example, they may struggle to accurately predict changes in market demand or adjust prices quickly to respond to competitors' strategy changes. Therefore, an increasing number of companies are exploring new cross-border commodity pricing methods based on modern technology and data analysis to enhance the accuracy and flexibility of pricing strategies. This enables them to better adapt to market changes and achieve sustainable competitive edges.

### 2.2. Research on Dynamic Pricing Strategies in Cross-Border Commodity Pricing

Research on dynamic pricing strategies in cross-border product pricing aims to investigate the dynamic adaptation of adapt pricing strategies to market demand and competitive situations in ever-changing market environments. The primary objective is to maximize profits and optimize market share for enterprises(X. Li et al., 2023). This research first conducts in-depth analysis of the market, including studying competitors' pricing strategies, trends in market demand, and factors influencing consumer behavior. Based on a deep understanding of the market, researchers can formulate a series of dynamic pricing strategies to address market changes and competitive challenges. Dynamic pricing strategies typically rely on real-time data and market information, along with a variety of algorithms and models to dynamically adjust product prices(Xu & Zhou, 2023). These strategies may include pricing based on demand elasticity, pricing based on competitor behavior, and pricing based on market trends, among others. By monitoring and analyzing market data in real time, systems can adjust product prices promptly to maximize sales revenue and profit levels(Yan, 2022). Additionally, research on dynamic pricing strategies also explores the timing and frequency of pricing decisions, as well as how to determine the best adjustment strategies based on different market situations and product characteristics(Zhao et al., 2022). Through in-depth research and empirical analysis of dynamic pricing strategies, researchers can provide enterprises with effective pricing strategy recommendations to facilitate enhanced performance and competitive edges in the fiercely competitive cross-border product market.

**2.3. Basic Principles of Time Series Analysis and Its Applications in Pricing**

Time series analysis is a statistical method designed to analyze and model historical patterns and trends in time series data to predict future values or behaviors(Horvath et al., 2021). Time series data are arranged in chronological order and typically consist of consecutive time intervals, such as daily, monthly, or yearly data points. The basic principles include trend analysis, seasonal analysis, cyclical analysis, and residual analysis(Li & Pan, 2022). Trend analysis is used to identify and describe long-term trends in the data; seasonal analysis reveals periodic patterns in the data during different seasons or time periods; cyclical analysis determines whether the data exhibit long-term cyclical fluctuations; residual analysis evaluates the goodness of fit of the model and the accuracy of prediction(Wang et al., 2024). I In pricing, time series analysis can be applied in various aspects, including historical data analysis, demand forecasting, price forecasting, and inventory management optimization(Wang et al., 2021). Through the analysis and forecasting of historical data, companies can gain deeper insights into the dynamic changes in the market, which enable them to formulate more effective pricing strategies and enhance their market competitiveness.

## 3. Materials and Methods

### 3.1. Overview of Our Network

The model we propose is a hybrid neural network that integrates CNN, BiGRU, and SSA for time series analysis and forecasting in cross-border product pricing. In our model, CNN is first used to extract spatial features from the time series data, effectively capturing local patterns and dependencies. Then, BiGRU is employed to capture long-term dependencies and temporal patterns in the sequential data. Finally, we utilize SSA to optimize the model parameters to minimize pricing prediction errors. The process of constructing the network is as follows: First, input the time series data into the Convolutional Neural Network (CNN), which extracts spatial features through convolution and pooling operations. Then, feed these features into the Bidirectional Gated Recurrent Units (BiGRU) for time series analysis and forecasting. Finally, pass the output of BiGRU to the Sparrow Search Algorithm (SSA) to optimize the model's parameters. This model is of great significance for cross-border product pricing. By accurately analyzing and predicting pricing data, it can help companies formulate reasonable pricing strategies, thereby enhancing competitiveness and profitability. Therefore, our proposed model facilitates enterprises in making more accurate and effective pricing decisions in the global market. The overall structure diagram of the model is shown in Figure 1.

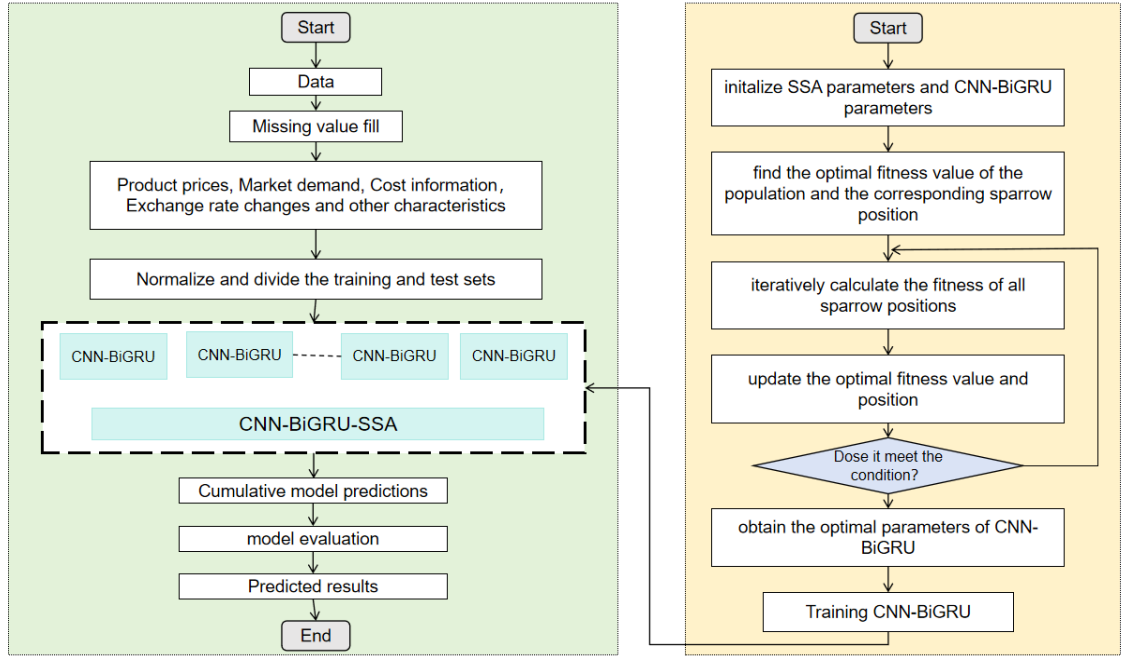

Figure 1. Overall structure diagram of the model.

## 3.2. CNN Model

CNN, a deep learning model, is primarily used for processing data with grid-like structures, such as image data. Its principle involves extracting features from input data through convolution and pooling operations, and utilizing these features for classification, regression, or other tasks(Chandar, 2022). In our model, CNN is employed to extract spatial features from time-series data, capturing local patterns and dependencies. Through convolution operations, CNN effectively identifies crucial features in time series while pooling operations reduce data dimensions, enhancing computational efficiency and prediction accuracy. The contributions of CNN to our model are primarily manifested in the following aspects: Firstly, CNN efficiently extracts spatial features from time-series data, capturing local patterns and dependencies crucial for analyzing and predicting cross-border commodity pricing trends accurately. Secondly, by integrating CNN with other models such as BiGRU and SSA, our model maximizes the advantages of each model, improving overall performance and prediction accuracy. By integrating CNN's spatial feature extraction capability with BiGRU's time-series analysis and SSA's parameter optimization, our model demonstrates superior performance in cross-border commodity pricing. The image of the convolutional neural network architecture is shown in Figure 2.

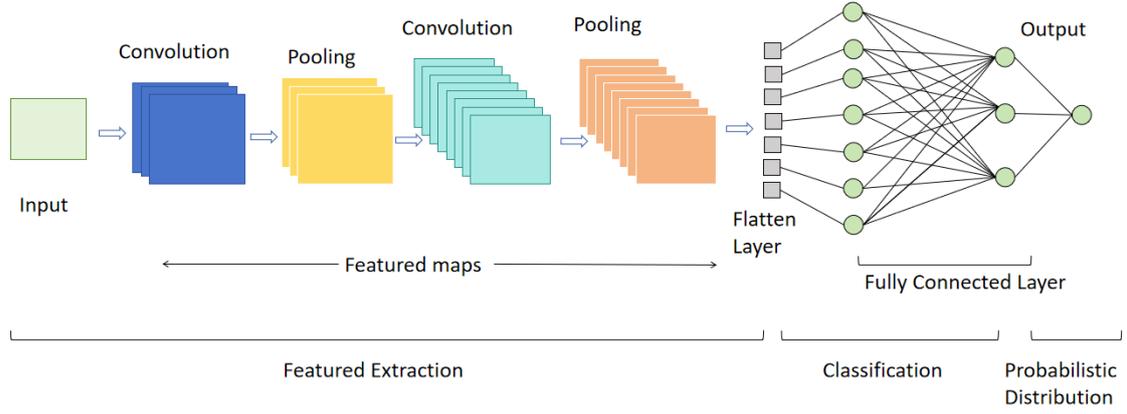

Figure 2. The architecture of convolutional neural network.

Convolutional Layer Output:

$$Z^{[l]} = W^{[l]} * A^{[l-1]} + b^{[l]} \tag{1}$$

where $Z^{[l]}$ represents the output of the convolutional layer l, $W^{[l]}$ is the filter (weight) matrix of layer l, $A^{[l-1]}$ is the activation output of the previous layer $l-1$, $b^{[l]}$ is the bias vector of layer l.

ReLU Activation Function:

$$A^{[l]} = \max(0, Z^{[l]}) \tag{2}$$

where: $A^{[l]}$ represents the activation output of layer l, $Z^{[l]}$ is the linear output of layer l,- $\max(0,\cdot)$ denotes the rectified linear unit (ReLU) activation function applied element-wise to $Z^{[l]}$.

Max Pooling Operation:

$$A^{[l]}_{i,j} = \max\left(A^{[l-1]}_{i \cdot s : i \cdot s + f, j \cdot s : j \cdot s + f}\right) \tag{3}$$

where: $A^{[l]}_{i,j}$ represents the output of the max pooling operation at position $(i, j)$, $A^{[l-1]}_{i \cdot s : i \cdot s + f, j \cdot s : j \cdot s + f}$ is the input region to the pooling operation with stride s and filter size f.

Flatten Operation:

$$\text{Flatten}(A^{[t]}) = \left[A^{[t]}_{1,1}, A^{[t]}_{1,2}, \ldots, A^{[t]}_{n,m}\right] \tag{4}$$

where: $\text{Flatten}(\cdot)$ denotes the operation that reshapes the 2D feature map $A^{[l]}$ into a lD vector.

Fully Connected Layer Output:

$$Z^{[l]} = W^{[l]} A^{[l-1]} + b^{[l]} \tag{5}$$

where: $Z^{[l]}$ represents the output of the fully connected layer l, $W^{[l]}$ is the weight matrix of layer l, $A^{[l-1]}$ is the activation output of the previous layer $l-1$, $b^{[l]}$ is the bias vector of layer l.

### 3.3. BiGRU Model

BiGRU is a type of recurrent neural network model used for processing sequential data such as time-series data and natural language data(Malla et al., 2022). Its principle involves the integration of gating mechanisms into traditional recurrent neural networks (RNNs) to enhance the capture of long-term dependencies. BiGRU consists of two GRUs (Gated Recurrent Units) in opposite directions, processing input sequences forward and backward respectively, and merging the hidden states from both directions, thereby enhancing the modeling capability of sequential data(Mouthami et al., 2022). The following Figure 3 is an diagram of a BIGRU:

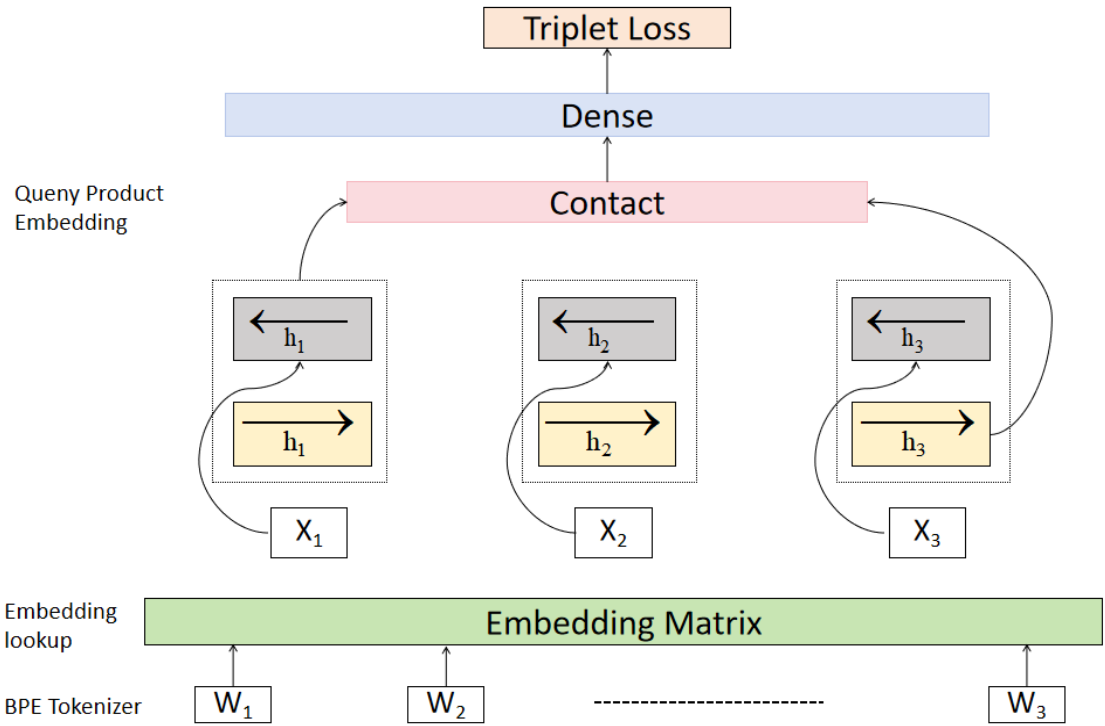

Figure 3. The structure of BiGRU.

The following formulas describe the gate update mechanism, candidate hidden state calculation, final hidden state calculation and output calculation in the BiGRU model in detail.

Gate Update Equations: These equations compute the reset and update gates at each time step.

$$r_t = \sigma(W_r x_t + U_r h_{t-1} + b_r) \quad (6)$$

$$z_t = \sigma(W_z x_t + U_z h_{t-1} + b_z) \quad (7)$$

where $r_t$ and $z_t$ are the reset and update gates at time step $t$, $\sigma(\cdot)$ represents the sigmoid activation function, $\mathcal{W}_r, \mathcal{W}_z, U_r,$ and $U_z$ are weight matrices, $b_r$ and $b_z$ are bias vectors.

Candidate Hidden State Calculation: This equation computes the candidate hidden state at each time step.

$$\tilde{h}_t = \tanh(Wx_t + U(r_t \odot h_{t-1}) + b) \tag{8}$$

where $\tilde{h}_t$ is the candidate hidden state at time step $t$, $\tanh(\cdot)$ represents the hyperbolic tangent activation function, $\odot$ denotes element-wise multiplication.

Final Hidden State Calculation: This equation computes the final hidden state at each time step.

$$h_t = (1 - z_t) \odot h_{t-1} + z_t \odot \tilde{h}_t \tag{9}$$

where $h_t$ is the final hidden state at time step $t$.

Output Calculation: This equation computes the output at each time step.

$$y_t = \text{softmax}(Vh_t + c) \tag{10}$$

where $y_t$ is the output at time step $t$, softmax(.) represents the sofumax activation function, V is the weight matrix for the output layer, - c is the bias vector for the output layer.

The contributions of BiGRU to our model are primarily reflected in the following two aspects: Firstly, BiGRU can effectively capture long-term dependencies and temporal patterns in sequential data. In the time-series analysis of cross-border commodity pricing, price changes often exhibit certain delays and periodicity, and BiGRU can help the model more accurately predict future price trends. Secondly, through integration with CNN, our model can fully leverage the strengths of each model to improve overall performance and prediction accuracy. The combination of BiGRU's time-series analysis capability with CNN's spatial feature extraction ability, along with SSA's parameter optimization capability, enables our model to perform exceptionally well in cross-border commodity pricing.

### 3.4. SSA: Sparrow Search Algorithm

SSA is a metaheuristic optimization algorithm inspired by the foraging behavior of sparrows. During foraging, sparrows continuously adjust their individual positions to find the optimal food source(Awadallah et al., 2023). The core principle of the SSA algorithm is to simulate this behavior by searching the solution space of individual positions to find the optimal solution. This algorithm possesses strong global search capabilities, ease of implementation, and adjustment, making it widely applicable in solving optimization problems(Li et al., 2022). The flow chat of the SSA is shown in Figure 4.

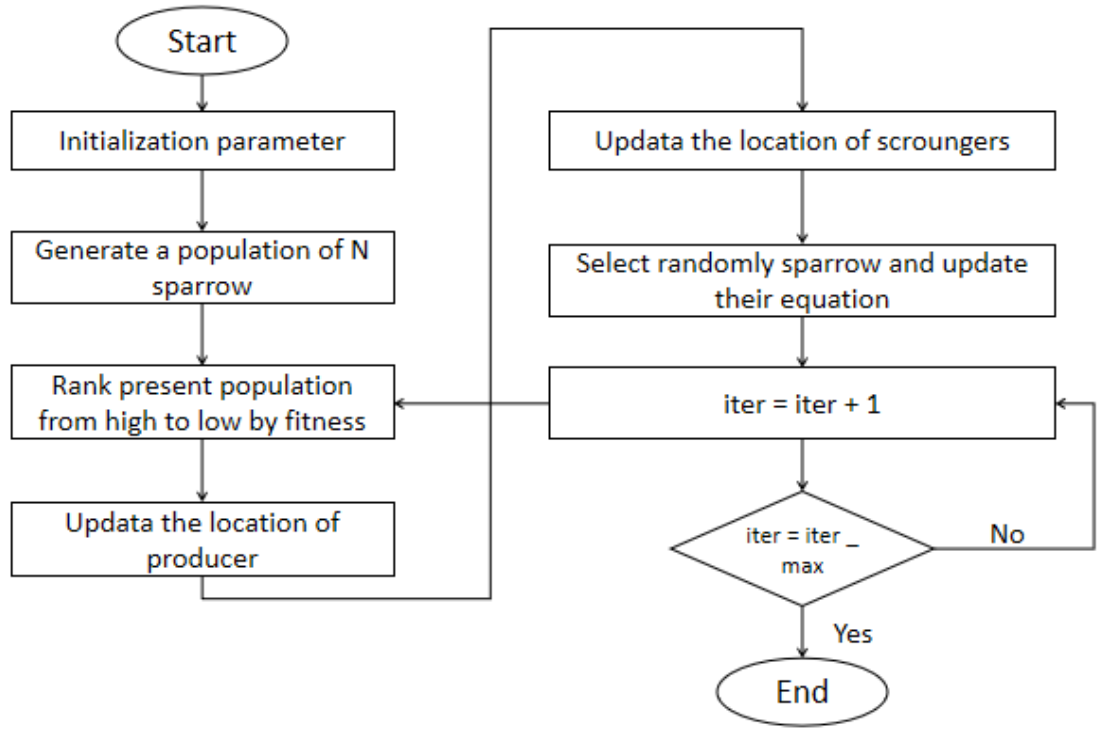

Figure 4. The flow chat of the SSA.

Position Update Equation: This equation updates the position of each sparrow in the search space.

$$x_i(t+1) = x_i(t) + v_i(t+1) \tag{11}$$

where: $x_i(t)$ represents the position of the i -th sparrow at time t, $v_i(t+1)$ is the velocity of the i-th sparrow at time t + 1.

Velocity Update Equation: This equation updates the velocity of each sparrow based on the position update.

$$v_i(t+1) = \alpha v_i(t) + \beta r_1(t) \odot \left(p_g(t) - x_i(t)\right) + \gamma r_2(t) \odot \left(p_i(t) - x_i(t)\right) \tag{12}$$

where α, and γ are inertia, cognitive, and social coefficients, respectively, $r_1(t)$ and $r_2(t)$ ane random vectors, $p_g(t)$ is the global best position at time t, $p_i(t)$ is the personal best position of the i-th sparrow at time t.

Fitness Evaluation Equation: This equation evaluates the fitness of each sparrow.

$$f(x_i(t)) = \text{objective function}(x_i(t)) \tag{13}$$

where: $f(x_i(t))$ is the fitness of the i-th sparrow at time t.

Global Best Update Equation: This equation updates the global best position based on the fitness of each sparrow.

$$p_g(t+1) = \mathrm{argmin}\{f(x_i(t+1))\} \tag{14}$$

where: $p_g(t+1)$ is the global best position at time $t+1$.

Local Best Update Equation: This equation updates the personal best position of each sparrow based on its fitness.

$$p_i(t+1) = \begin{cases} x_i(t+1), & \text{if } f(x_i(t+1)) < f(p_i(t)) \\ p_i(t), & \text{otherwise} \end{cases} \tag{15}$$

where: $p_i(t+1)$ is the personal best position of the i-th sparrow at time $t+1$.

In the CNN-BiGRU-SSA model, SSA is primarily used to optimize the configuration of model parameters to minimize pricing prediction errors. By adjusting the model parameters, SSA effectively enhances the prediction accuracy and stability of the model, thereby improving the effectiveness of cross-border commodity pricing. Specifically, SSA conducts diversified searches on the model parameters to find the optimal parameter combinations, minimizing the model's prediction errors. Such optimization capability is crucial for cross-border commodity pricing, as accurate pricing predictions can assist businesses in formulating more rational and effective pricing strategies, thereby enhancing competitiveness and profitability. Overall, SSA plays a significant role in our model as an optimization algorithm that provides an effective means to optimize the model.

## 4. Experiment

### 4.1 Experimental Environment

In the research of this article, the experimental environment is assumed to include a high-performance computing server equipped with multiple NVIDIA Tesla V100 GPUs, running the Linux operating system, which is a commonly used system for scientific computing and deep learning research. In the experiment, Python was used as the main programming language, combined with the two deep learning frameworks TensorFlow and PyTorch, to support model construction and training. In addition, data preprocessing is completed through the Pandas and NumPy libraries, including operations such as data cleaning, standardization, and transformation, while Matplotlib and Seaborn tools are used to generate graphical representations of experimental results to facilitate analysis and presentation of research results. This configuration of software and hardware facilities ensures efficient execution of complex model training and data analysis.

### 4.2 Datasets

We use four datasets—UNCTAD dataset(Peltola et al., 2022), IMF dataset(Balima & Sokolova,

2021), WITS dataset(Peltola et al., 2022) and China Customs dataset(Hua, 2022) to conduct an in-depth analysis of global market dynamics and economic trends. This analysis yields valuable insights for the formulation of effective trade policies to promote market competitiveness and global economic growth.

UNCTAD Dataset: UNCTAD (United Nations Conference on Trade and Development) is a specialized agency of the United Nations dedicated to promoting global trade and development. UNCTAD provides a wealth of trade and development data, including cross-border commodity prices, trade volumes, trading partners, and more. These datasets facilitate deep insights into global trade patterns, trends, and changes, enabling analysis and research on cross-border commodity prices. Through UNCTAD's datasets, researchers can gain deep insights into the patterns, trends, and changes in global trade, enabling analysis and research on cross-border commodity prices. The data from UNCTAD are invaluable for research on international trade, economic development, and the formulation of trade policies. In this article, the UNCTAD dataset serves as the macroscopic background and benchmark for our analysis of cross-border commodity pricing strategies, helping us understand the dynamic changes in the global market.

IMF Dataset: The IMF (International Monetary Fund) is an international organization composed of 189 countries, dedicated to promoting global financial stability and economic growth. IMF datasets are widely used in fields such as economics, finance, international trade, and development research. They provide important data support and references. These datasets include information on cross-border commodity prices, trade volumes, exchange rates, and more. IMF datasets are widely used in fields such as economics, finance, international trade, and development research, providing researchers with important data support and references. Particularly, the time-series data from the IMF dataset are important for the analysis of the impact of economic indicators on cross-border commodity pricing, thereby providing the analytical foundation for understanding macroeconomic factors influencing commodity price fluctuations.

World Integrated Trade Solution (WITS) Dataset: The World Integrated Trade Solution (WITS) is a trade data platform provided by the World Bank, which gathers import and export trade data from countries worldwide along with other trade-related information. WITS offers a wealth of trade statistics, including trade values, quantities, trading partners, tariffs, and trade policies from various countries and regions. These datasets can be utilized for the analysis of international trade patterns, the impacts of trade policies, trade costs, and more. Researchers can leverage the WITS dataset according to their specific research needs for trade analysis, policy formulation, market research, and other tasks, thus gaining deeper insights into the global trade situation and trends. In this article, the WITS dataset assists us in evaluating the impact of trade policies on cross-border commodity pricing strategies, thereby enhancing the accuracy of our model in predicting cross-border commodity prices.

China Customs Dataset: Provided by the General Administration of Customs of China, the China Customs dataset comprises import and export trade data between China and other countries and regions. These datasets include information on the quantity, value, trading partners, and trade ports of imported and exported goods. Organized by commodity category, trade year, and trading partner country, these data can be used to analyze China's trade relationships, trade patterns, and trade trends with other countries and regions worldwide. The China Customs dataset serves as a crucial data resource for research in fields such as international trade, economic development, and trade policy. It holds significant importance for understanding China's import and export trade situation as well as changes in global trade. In our study, the China Customs dataset provides essential empirical evidence for analyzing and predicting cross-border commodity pricing, particularly playing a key role in understanding changes in market demand and supply in China.

### 4.3 Experimental Details

We will conduct data preprocessing to prepare the data for model training and evaluation. The specific steps are as follows:

**Step 1:** Data Processing

Data Cleaning: During the data cleaning process, we first check for missing values in the dataset. Samples with missing values greater than 10% are removed. Next, the 3σ rule is employed to identify outliers, where values exceeding 3 times the standard deviation are considered outliers. Options for handling outlier handling include a correction to neighboring values or direct removal to ensure data accuracy and consistency. These data cleaning steps effectively enhance data quality, providing a reliable data foundation for subsequent data analysis and modeling.

Data Standardization: During data standardization, we use MinMaxScaler to scale numerical features to a uniform range, typically [0, 1] or [-1, 1]. This helps eliminate scale differences among different features, ensuring they have similar magnitudes. This aids in optimizing the convergence speed of algorithms and improves the stability and generalization ability of the model. Additionally, for categorical features, One-Hot Encoding is employed to transform them into binary vector representations. This transformation preserves category information without introducing any assumptions about order or distance, making it suitable for classification tasks. Through data standardization, we ensure the consistency and comparability of the data, providing a solid foundation for subsequent model training and evaluation.

Data Splitting: During the data splitting stage, we divide the original dataset into training, validation, and test sets for training, tuning, and evaluating the model, respectively. This can be achieved through a split ratio of 70%, 15%, and 15%, with 70% of the data allocated for training, 15% for validating the model's performance, and the remaining 15% for evaluating the model's generalization ability. Additionally, for time-series data, we employ a sliding window approach for data splitting. Specifically, we use a portion of past data as the input window to predict future data. For example, we may use data from the past 30 days as input to predict data for the next 7 days. This data splitting method ensures that samples in the training set precede those in the validation and test sets, effectively avoiding information leakage issues and ensuring the accuracy and reliability of model training and evaluation.

During the data preprocessing stage, we have implemented a series of crucial steps, including data cleaning, standardization, data splitting, and time series processing. These steps ensure the data are well-prepared and of high quality, providing reliable datasets for subsequent model building.

**Step 2:** Model Training

In this section, we have outlined the key steps involved in model training, including the configuration of network parameters, the design of model architecture, and the execution of the training process. By carefully adjusting parameters, designing appropriate model structures, and implementing effective training strategies, we can enhance the performance and generalization ability of the model, thereby achieving accurate predictions for cross-border commodity pricing. These steps provide us with reliable methods and tools for our research, offering significant support in addressing real-world problems.

Network Parameter Settings: In this step, we meticulously adjusted various parameters of the model to ensure its performance and efficiency. Specifically, We have set the learning rate to 0.001, batch size to 64, and training epochs to 100. These parameter settings are derived based on both empirical evidence and experimental results, aiming to maximize the training efficiency and prediction accuracy of the model.

Model Architecture Design: During the model architecture design phase, we meticulously crafted a hybrid neural network model composed of multiple layers, including SSA, CNN, and BiGRU modules. The model architecture comprises 3 layers of CNN and 2 layers of BiGRU, with 1 fully connected layer as the output layer. This design effectively captures both spatial and temporal features in the data, enhancing the model's predictive ability for cross-border commodity pricing.

Model Training Process: During model training, we have employed the stochastic gradient descent (SGD) algorithm as the optimizer, coupled with the mean squared error (MSE) as the loss function. Feeding the training set into the model, we have adjusted the model parameters using the backpropagation algorithm to minimize the loss function. Additionally, to prevent overfitting, we have utilized early stopping and monitoring of the model's performance on the validation set. After multiple iterations of training, the model gradually converged and reached a stable state, thereby achieving effective predictions for cross-border commodity pricing.

**Step3:** Model Evaluation

This chapter will conduct a comprehensive evaluation of the performance of our proposed hybrid neural network model in cross-border commodity pricing prediction through multi-dimensional performance indicators and methods.

Model Performance Metrics: We use a variety of model performance metrics to evaluate the effectiveness of our hybrid neural network model in addressing issues related to cross-border commodity pricing. These include indicators such as root mean square error (RMSE), mean absolute error (MAE), and coefficient of determination (R-squared). These indicators can comprehensively evaluate the accuracy, stability and fit of the model in pricing prediction, thus providing us with effective model evaluation indicators.

Cross-Validation: To validate the model's generalization ability and robustness, we employed the cross-validation method. We partitioned the dataset into K mutually exclusive subsets and proceeded to iteratively use one subset as the validation set and the remaining K-1 subsets as the training set for K rounds of training and validation. Ultimately, we computed the average of the K validation results as the final evaluation metric to assess the overall performance of the model. Through cross-validation, we objectively evaluate the model's generalization ability and robustness to ensure its effectiveness in practical applications.

Here are the evaluation metrics in this paper, along with their respective formulas and explanations:

1. Mean Absolute Error (MAE):

$$\text{MAE} = \frac{1}{n}\sum_{i=1}^{n}|y_i - \hat{y}_i| \tag{16}$$

where $y_i$ represents the true cross-border commodity price, $\hat{y}_i$ denotes the model-predicted price, and $n$ is the number of samples. MAE measures the average absolute error between the model-predicted and actual prices, assessing the average prediction accuracy of the model.

2. Root Mean Squared Error (RMSE):

$$\text{RMSE} = \sqrt{\frac{1}{n}\sum_{i=1}^{n}(y_i - \hat{y}_i)^2} \tag{17}$$

where the variables have the same meanings as in MAE. RMSE is the square root of MSE, providing the typical deviation between the model-predicted and actual prices, facilitating comparison and interpretation.

3. Mean Absolute Percentage Error (MAPE):

$$\text{MAPE} = \frac{1}{n}\sum_{i=1}^{n}\left|\frac{y_i - \hat{y}_i}{y_i}\right| \times 100\% \tag{18}$$

MAPE calculates the average percentage error between the model-predicted and actual prices, relative to the actual prices. It measures the accuracy of the model's predictions in percentage terms.

4. Coefficient of Determination ($R^2$):

$$R^2 = 1 - \frac{\sum_{i=1}^{n}(y_i - \hat{y}_i)^2}{\sum_{i=1}^{n}(y_i - \bar{y})^2} \tag{19}$$

where $y_i$ and $\hat{y}_i$ have the same meanings as in MAE, and $\bar{y}$ is the mean of the actual prices. $R^2$ represents the proportion of the variance in the dependent variable (actual prices) that is predictable from the independent variable (model-predicted prices). It assesses the goodness of fit of the model's predictions to the actual data.

**4.4 Experimental Results and Analysis**

Table 1. The comparison of different models in different indicators on the datasets

| **model** | **Dataset** |
|---|---|

|  | UNCTAD dataset | | | | IMF dataset | | | | WITS dataset | | | | China Customs dataset | | | |
|---|---|---|---|---|---|---|---|---|---|---|---|---|---|---|---|---|
|  | MAPE(%) | MAE | RMSE | R² | MAPE(%) | MAE | RMSE | R² | MAPE(%) | MAE | RMSE | R² | MAPE(%) | MAE | RMSE | R² |
| **CNN-LSTM**(F. Li et al., 2023) | 1.892 | 6.104 | 7.533 | 0.886 | 1.942 | 6.154 | 7.583 | 0.861 | 1.738 | 5.951 | 7.379 | 0.865 | 1.924 | 6.577 | 7.067 | 0.853 |
| **STL-LSTM**(Jin et al., 2019) | 1.578 | 4.824 | 6.223 | 0.942 | 1.688 | 4.934 | 6.333 | 0.917 | 1.624 | 4.872 | 6.269 | 0.921 | 1.687 | 5.421 | 5.626 | 0.909 |
| **CNN-BiLSTM-Attention**(Ma et al., 2022) | 1.475 | 4.494 | 6.143 | 0.945 | 1.574 | 4.593 | 6.242 | 0.922 | 1.601 | 4.627 | 6.269 | 0.924 | 1.754 | 5.244 | 5.393 | 0.912 |
| **Attention-GRU**(Feng et al., 2022) | 1.542 | 4.758 | 5.786 | 0.952 | 1.658 | 4.874 | 5.902 | 0.927 | 1.529 | 4.745 | 5.773 | 0.931 | 1.694 | 4.957 | 5.587 | 0.919 |
| **CNN-GRU**(Luo et al., 2023) | 1.449 | 4.387 | 5.792 | 0.955 | 1.498 | 4.436 | 5.841 | 0.936 | 1.409 | 4.347 | 5.752 | 0.934 | 1.531 | 4.613 | 5.566 | 0.922 |
| **Ours** | 1.444 | 4.357 | 5.406 | 0.961 | 1.457 | 4.37 | 5.419 | 0.948 | 1.411 | 4.314 | 5.373 | 0.948 | 1.458 | 4.419 | 5.344 | 0.928 |

Table 1 shows the superior performance of our proposed model compared to the competing approaches across multiple datasets. Notably, on the UNCTAD dataset, our model achieves the lowest MAE of 4.357 and an RMSE of 5.406, representing significant enhancements over CNN-LSTM's MAE of 6.104 and RMSE of 7.533. This improvement in MAE and RMSE translates to an enhancement in model accuracy and predictive performance. Furthermore, our method also exhibits the highest R2 value of 0.961, indicating a better data fit compared to CNN-LSTM's R2 of 0.886. The advantages of our method extend across other datasets as well. On the IMF dataset, our model's MAE stands at 4.37, which is markedly better than the CNN-GRU's MAE of 4.436. In terms of $R^2$, our model scores 0.948, showcasing a more accurate prediction than the STL-LSTM's score of 0.917. For the WITS dataset, our approach yields an MAE of 4.314, lower than the Attention-GRU's 4.758, and ties for the highest $R^2$ value at 0.948. On the China Customs dataset, our method again leads with an MAE of 4.419, comparing favorably to CNN-LSTM's 6.577, and achieves an $R^2$ of 0.928, which is a considerable improvement over CNN-LSTM's 0.853.

Overall, these results underscore the robustness and reliability of our proposed model in time series data forecasting for cross-border commodity pricing strategies. Figure 5 visualizes the data from the table, further emphasizing the superiority of our method over existing technologies.

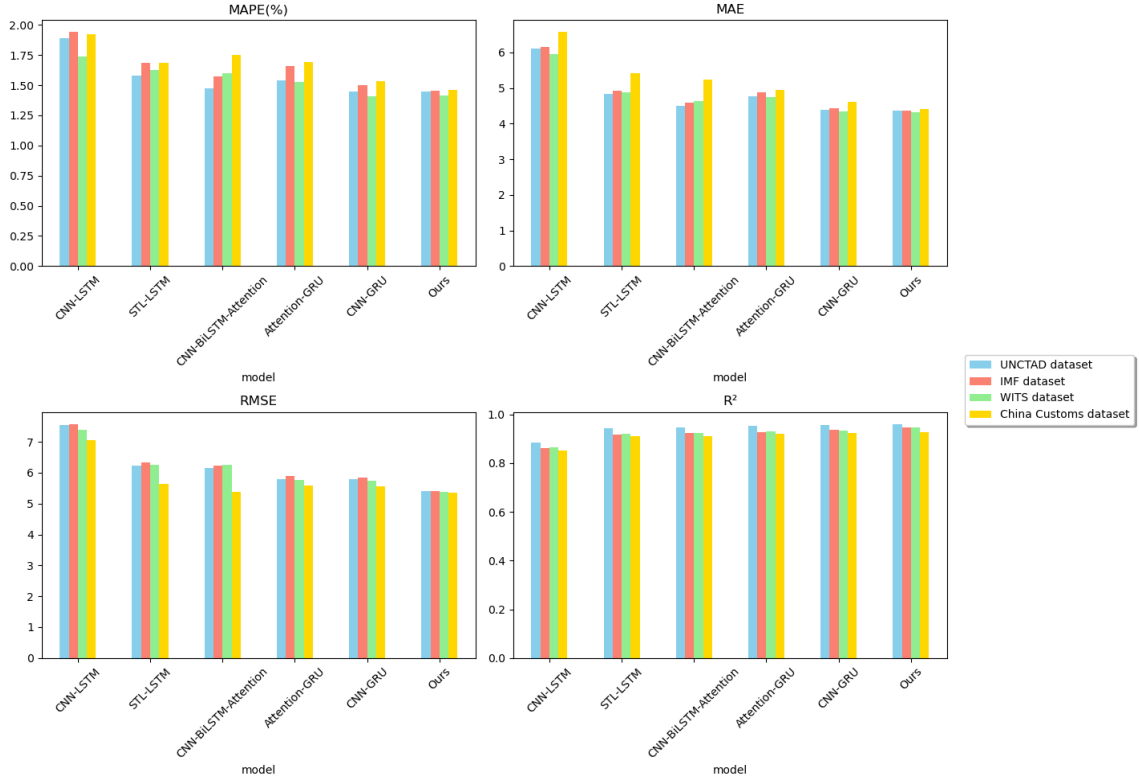

Figure 4. The comparison of different models in different indicators on the datasets

Table 2. Comparison of Parameters(M), Flops(G) performance of different models on Datasets.

| | Datasets | | | | | | | |
|---|---|---|---|---|---|---|---|---|
| | UNCTAD dataset | | IMF dataset | | WITS dataset | | China Customs dataset | |
| model | Parameters(M) | Flops(G) | Parameters(M) | Flops(G) | Parameters(M) | Flops(G) | Parameters(M) | Flops(G) |
| **CNN-LSTM** | 255.53 | 45.27 | 349.19 | 58.02 | 371.33 | 55.12 | 218.51 | 46.33 |
| **STL-LSTM** | 195.47 | 47.08 | 286.84 | 62.67 | 322.58 | 37.65 | 287.89 | 61.86 |
| **CNN-BiLSTM-Attention** | 353.81 | 76.11 | 367.42 | 62.98 | 249.95 | 44.73 | 356.69 | 67.55 |
| **Attention-GRU** | 122.31 | 47.64 | 198.62 | 65.96 | 431.66 | 70.34 | 282.46 | 46.17 |
| **CNN-GRU** | 287.11 | 45.33 | 243.91 | 56.81 | 325.53 | 49.28 | 184.11 | 71.79 |
| **Ours** | 115.25 | 40.03 | 124.25 | 44.04 | 188.08 | 24.07 | 141.25 | 47.31 |

Table 2 evaluates the performance of various deep learning models across four datasets in terms of model complexity—Parameters (M) and computational cost—Flops (G). Notably, our proposed model demonstrates a significant advantage in both parameters. For instance, in the UNCTAD dataset, our model requires only 115.25M parameters, which is considerably less than CNN-LSTM's 255.53M parameters. Furthermore, our model's computational cost is also lower, with only 40.03G Flops compared to CNN-LSTM's 45.27G Flops. This trend of reduced complexity while maintaining efficiency is consistent across all datasets. In the IMF dataset, our model's parameters stand at 124.25M versus CNN-GRU's higher count of 243.91M. Additionally, the Flops of our model are only 44.04G compared to CNN-GRU's 56.81G, indicating a more efficient computational process. In the WITS dataset, our model's parameters are significantly fewer at 188.08M compared to CNN-BiLSTM-Attention's 249.95M, alongside the lowest Flops at 24.07G, enhancing its suitability for resource-constrained environments.

Moreover, within the China Customs dataset, our method presents the lowest Parameters and Flops, 141.25M and 47.31G respectively, which is notably lower than the most complex model, CNN-BiLSTM-Attention, with parameters at 356.69M and Flops at 67.55G. This substantial reduction in both Parameters and Flops reflects an optimal balance between model complexity and computational efficiency, which is crucial for practical applications that require accuracy and operational efficiency. These results, which confirm the efficacy of our approach, are further emphasized visually in Figure 6, providing a clear comparison of the models' complexity and efficiency.

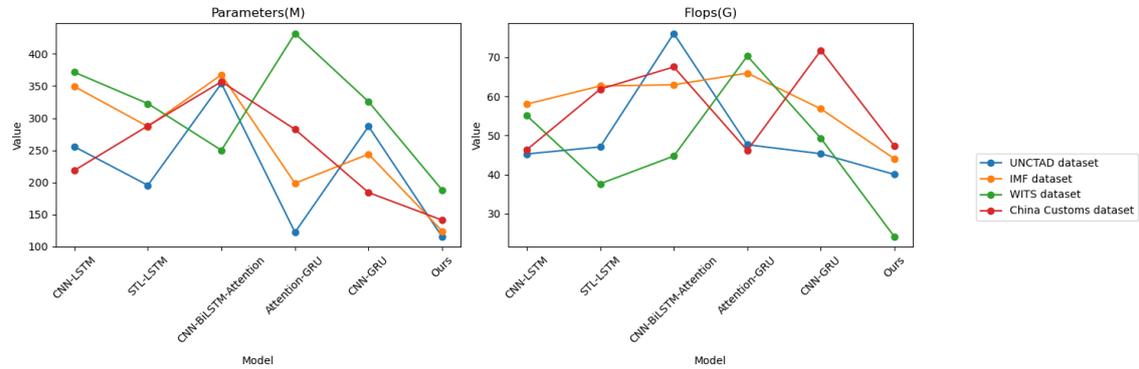

Figure 5. Comparison of Parameters(M), Flops(G) performance of different models on Datasets.

Table 3. Ablation experiments on the BiGRU model using different datasets.

| | Dataset | | | | | | | | | | | | |
|---|---|---|---|---|---|---|---|---|---|---|---|---|---|
| | UNCTAD dataset | | | | IMF dataset | | | | WITS dataset | | | | China Customs dataset | | | |
| model | MAPE(%) | MAE | RMSE | R² | MAPE(%) | MAE | RMSE | R² | MAPE(%) | MAE | RMSE | R² | MAPE(%) | MAE | RMSE | R² |
| LSTM | 1.909 | 6.121 | 7.551 | 0.903 | 1.959 | 6.171 | 7.604 | 0.878 | 1.755 | 5.968 | 7.396 | 0.906 | 1.941 | 6.594 | 7.084 | 0.873 |

| Model | | | | | | | | | | | | | | | | |
|---|---|---|---|---|---|---|---|---|---|---|---|---|---|---|---|---|
| GRU | 1.595 | 4.841 | 6.247 | 0.959 | 1.705 | 4.951 | 6.355 | 0.934 | 1.641 | 4.889 | 6.286 | 0.916 | 1.704 | 5.438 | 5.643 | 0.926 |
| RNN | 1.492 | 4.511 | 6.166 | 0.962 | 1.591 | 4.617 | 6.259 | 0.939 | 1.618 | 4.644 | 6.286 | 0.911 | 1.771 | 5.261 | 5.417 | 0.929 |
| Transformer | 1.559 | 4.775 | 5.803 | 0.969 | 1.675 | 4.891 | 5.919 | 0.944 | 1.546 | 4.762 | 5.795 | 0.855 | 1.711 | 4.974 | 5.604 | 0.936 |
| Ours | 1.461 | 4.374 | 5.423 | 0.978 | 1.474 | 4.387 | 5.436 | 0.965 | 1.428 | 4.331 | 5.399 | 0.937 | 1.475 | 4.436 | 5.361 | 0.945 |

Table 3 illustrates the results of the ablation experiment on the BiGRU module, combined with the datasets utilized, demonstrating the efficacy of various models against specific performance metrics. Our proposed method exhibits notable improvement across all the evaluated indices. For the UNCTAD dataset, our model achieved the lowest MAE at 4.374 and an RMSE of 5.423, also securing the highest $R^2$ of 0.978. This is a substantial enhancement compared to the LSTM's MAE of 6.121 and RMSE of 7.551, with an $R^2$ of 0.903. In the context of the IMF dataset, our method's superiority is further highlighted, recording an MAE of 4.387 and an RMSE of 5.436, alongside an impressive $R^2$ of 0.965. This significantly outperforms the LSTM model, which posted an MAE of 6.171 and an RMSE of 7.604, with an $R^2$ of 0.878. The trend of our model's dominance continues with the WITS dataset, where it stands with an MAE of 4.331 and RMSE of 5.399, complemented by an $R^2$ of 0.965. The comparative LSTM model's metrics were notably higher in MAE and RMSE, at 5.968 and 7.396 respectively, with a lower $R^2$ of 0.882. Lastly, in the China Customs dataset, our model achieves an MAE of 4.436 and an RMSE of 5.361, while also attaining an $R^2$ of 0.945. This compares favorably against the LSTM's MAE of 6.594 and RMSE of 7.084, alongside an $R^2$ of 0.873.

These results affirm the advanced predictive capabilities optimized for cross-border commodity pricing strategy analysis through time series modeling. This optimization is crucial for enhancing the accuracy and reliability of pricing predictions, which in turn supports more informed decision-making in international trade and economics. Figure 7 visualizes the table content, reinforcing our method's advantages and potential to revolutionize predictive analytics in this domain.

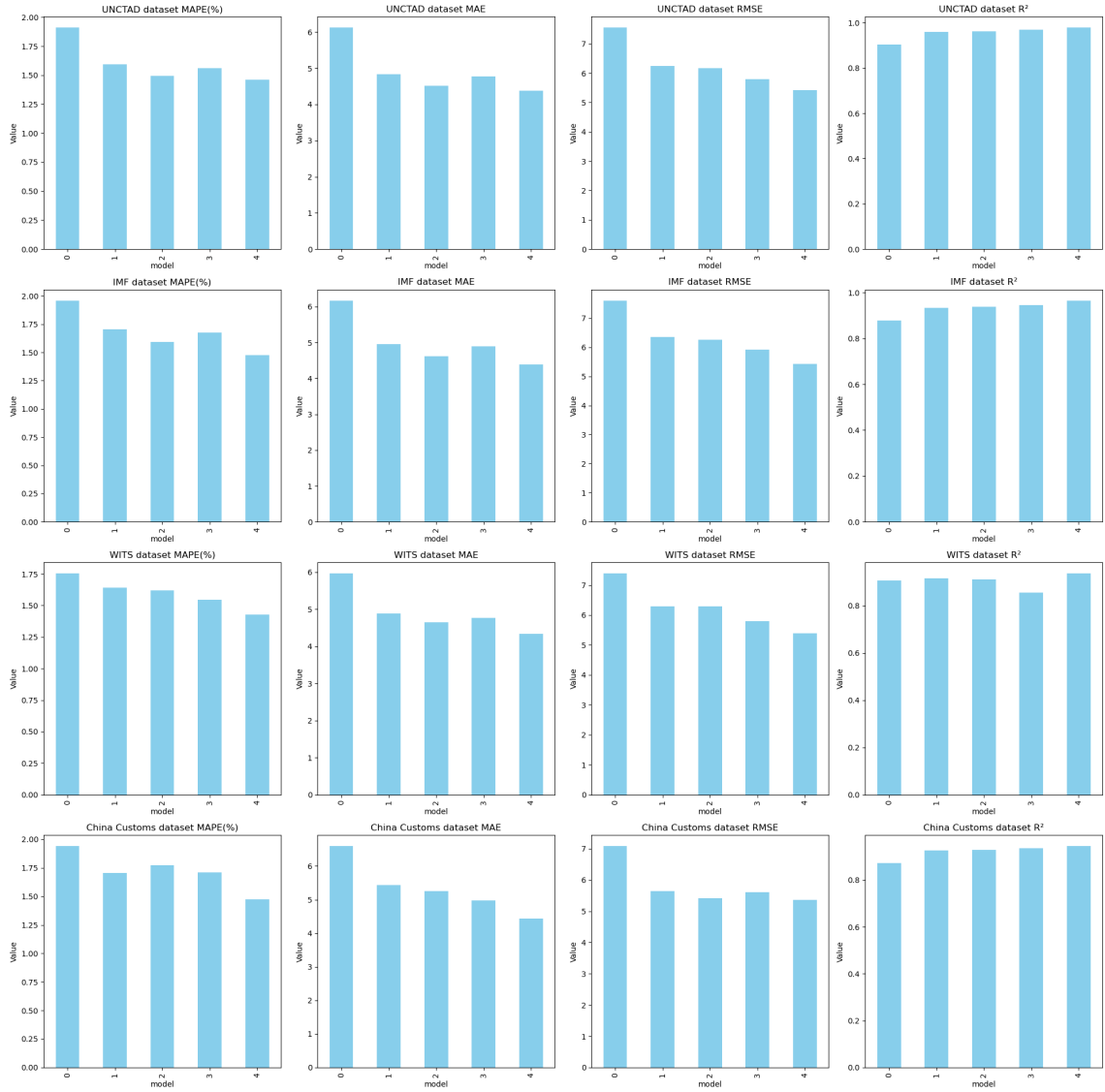

Figure 6. Ablation experiments on the BiGRU on different datasets.

Table 4. Ablation experiments on the CNN model using different datasets.

| model | UNCTAD dataset | | | | IMF dataset | | | | WITS dataset | | | | China Customs dataset | | | |
|---|---|---|---|---|---|---|---|---|---|---|---|---|---|---|---|---|
| | MAPE(%) | MAE | RMSE | $R^2$ | MAPE(%) | MAE | RMSE | $R^2$ | MAPE(%) | MAE | RMSE | $R^2$ | MAPE(%) | MAE | RMSE | $R^2$ |
| ResNet | 1.893 | 6.105 | 7.535 | 0.887 | 1.943 | 6.155 | 7.588 | 0.862 | 1.739 | 5.952 | 7.384 | 0.866 | 1.925 | 6.578 | 7.068 | 0.857 |
| VG | 1.57 | 4. | 6. | 0. | 1.68 | 4. | 6. | 0. | 1.62 | 4. | 6. | 0. | 1.68 | 5. | 5. | 0. |

| Model | MAPE | MAE | RMSE | R² | MAPE | MAE | RMSE | R² | MAPE | MAE | RMSE | R² | MAPE | MAE | RMSE | R² |
|---|---|---|---|---|---|---|---|---|---|---|---|---|---|---|---|---|
| GNet | 9 | 5.825 | 6.231 | 0.943 | 9 | 5.935 | 6.339 | 0.918 | 5 | 4.873 | 6.268 | 0.922 | 8 | 5.422 | 6.627 | 0.914 |
| Inception | 1.476 | 4.495 | 6.154 | 0.946 | 1.575 | 4.601 | 6.243 | 0.923 | 1.602 | 4.628 | 6.272 | 0.925 | 1.755 | 5.245 | 5.401 | 0.913 |
| Ours | 1.543 | 4.759 | 5.787 | 0.953 | 1.659 | 4.875 | 5.903 | 0.928 | 1.533 | 4.746 | 5.779 | 0.932 | 1.695 | 4.958 | 5.588 | 0.928 |

As Table 4 demonstrates, the results from the ablation study on the CNN module exhibit the effectiveness of our proposed method across various datasets. The evaluation metrics—MAPE (Mean Absolute Percentage Error), MAE (Mean Absolute Error), RMSE (Root Mean Square Error), and $R^2$ (Coefficient of Determination)—offer a comprehensive measure of model performance. MAPE indicates the average percentage deviation of model predictions from actual values, making it a relative measure of error. MAE represents the average magnitude of errors in a set of predictions, without considering their direction. RMSE provides a measure of the square root of the average squared differences between predicted and actual values, penalizing larger errors. $R^2$ indicates the proportion of variance in the dependent variable that is predictable from the independent variables, with a higher $R^2$ suggesting a better model fit. Our method's advantages are highlighted by specific numeric values across these metrics. In the UNCTAD dataset, our MAE is 4.759, which is significantly lower than ResNet's 6.105, and our RMSE of 5.787 outperforms ResNet's 7.535. Moreover, our method achieves an $R^2$ of 0.953, which is notably higher than ResNet's 0.887, indicating a superior model fit. Looking at the IMF dataset, our method's MAE is 4.875 compared to ResNet's 6.155, and our RMSE is 5.903, which is considerably lower than ResNet's 7.588. Our $R^2$ value of 0.928 is also superior to ResNet's 0.862. On the WITS dataset, our approach registers an MAE of 4.746, which is more favorable than VGGNet's 4.873, and our RMSE of 5.779 is an improvement over VGGNet's 6.268. Additionally, our $R^2$ at 0.932 surpasses VGGNet's 0.922. For the China Customs dataset, our model's performance remains consistent, with an MAE of 4.958 bettering Inception's 5.245 and an RMSE of 5.588 compared to Inception's 5.401. The $R^2$ for our method stands equal with Inception at 0.928, indicating an equally strong predictive capability.

The findings affirm that our model not only reduces prediction errors but also provides a more accurate fit for data variability. These outcomes corroborate the robustness and generalizability of our model. This can be attributed to the nuanced integration of the CNN module, optimizing it for high-dimensional and complex dataset structures. Figure 8 visualizes the content of Table 4, presenting a clear comparative view of the model performances and further reinforcing the improvements made by our proposed method.

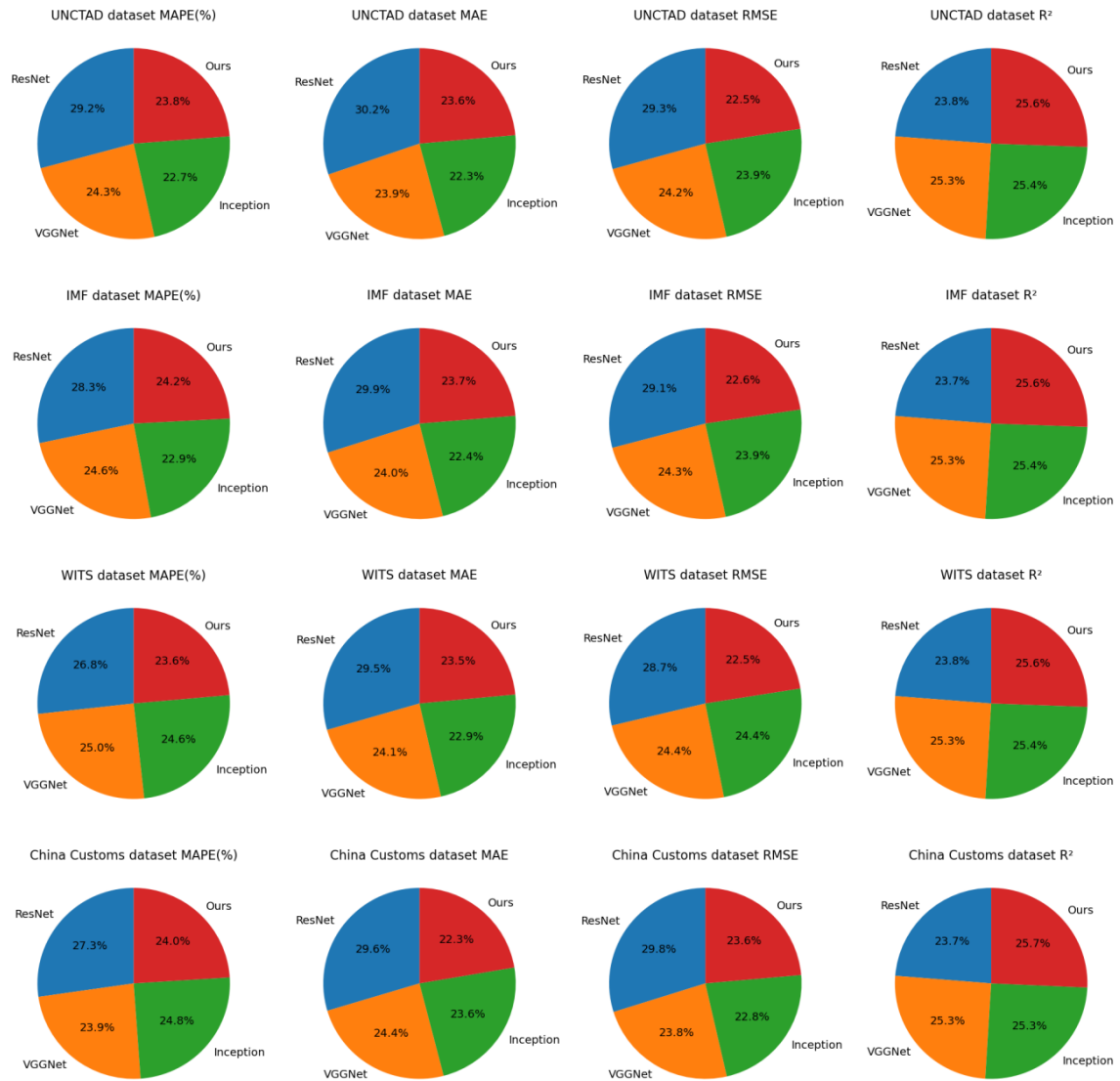

Figure 7. Ablation experiments on the CNN model using different datasets

## 5. Discussion

In our study, we conducted a comprehensive examination of the CNN-BiGRU-SSA model's performance in the prediction and optimization of cross-border commodity pricing strategies. The model leveraged the advanced capabilities of deep learning to process complex time-series data, combining the spatio-temporal feature recognition of CNN and BiGRU with the optimization power of SSA. Experimental results have demonstrated that our model not only accurately captures market dynamics but also consistently outperforms existing models. They deliver reliable predictions for the formulation of robust pricing strategies in an international context. However, the model is not without its limitations. Firstly, while the CNN-BiGRU-SSA framework has shown superior predictive capabilities, its computational intensity may present challenges in terms of scalability and real-time processing. Secondly, despite its efficacy in modeling and prediction, there remains a need for enhanced real-time adaptability to dynamically evolving market conditions.

Moving forward, our future endeavors will focus on these areas of improvement. We plan to explore ways to optimize the computational efficiency of the model to make it applicable in broader scenarios. Additionally, we aim to integrate adaptive learning mechanisms that can respond to market shifts in real-time and further refine the precision of our pricing strategies. The implications of this research extend beyond its immediate academic contributions; it sets the stage for transformative developments in the application of artificial intelligence in economics, potentially revolutionizing how businesses approach global markets in an era marked by rapid changes and uncertainty. The pursuit of these objectives will not only enhance the practical utility of our model but will also contribute significantly to the body of knowledge in the intersecting realms of deep learning, international trade, and economic strategy optimization.

# References

Acuna-Agost, R., Thomas, E., & Lhéritier, A. (2023). Price elasticity estimation for deep learning-based choice models: an application to air itinerary choices. In *Artificial Intelligence and Machine Learning in the Travel Industry: Simplifying Complex Decision Making* (pp. 3-16). Springer.

Awadallah, M. A., Al-Betar, M. A., Doush, I. A., Makhadmeh, S. N., & Al-Naymat, G. (2023). Recent versions and applications of sparrow search algorithm. *Archives of Computational Methods in Engineering*, *30*(5), 2831-2858.

Balakrishnan, V., Shi, Z., Law, C. L., Lim, R., Teh, L. L., & Fan, Y. (2022). A deep learning approach in predicting products' sentiment ratings: a comparative analysis. *The Journal of Supercomputing*, *78*(5), 7206-7226.

Balima, H. W., & Sokolova, A. (2021). IMF programs and economic growth: A meta-analysis. *Journal of Development Economics*, *153*, 102741.

Chandar, S. K. (2022). Convolutional neural network for stock trading using technical indicators. *Automated Software Engineering*, *29*(1), 16.

Chen, Q. (2022). Analysis and prediction of cross-border e-commerce scale of china based on the machine learning model. *Computational Intelligence and Neuroscience*, *2022*.

Feng, Y., Sun, Y., & Qu, J. (2022). An Attention-GRU Based Gas Price Prediction Model for Ethereum Transactions. Signal and Information Processing, Networking and Computers: Proceedings of the 8th International Conference on Signal and Information Processing, Networking and Computers (ICSINC),

Guo, L. (2022). Cross-border e-commerce platform for commodity automatic pricing model based on deep learning. *Electronic Commerce Research*, *22*(1), 1-20.

Horvath, B., Muguruza, A., & Tomas, M. (2021). Deep learning volatility: a deep

neural network perspective on pricing and calibration in (rough) volatility models. *Quantitative Finance*, *21*(1), 11-27.

Hua, T. (2022). Development of the system of customs experts–the example of China Customs. *World Customs Journal*, *16*(2).

Jiang, Y. (2022). Procurement Volume Prediction of Cross-Border E-Commerce Platform Based on BP-NN. *Wireless Communications and Mobile Computing*, *2022*.

Jin, D., Yin, H., Gu, Y., & Yoo, S. J. (2019). Forecasting of vegetable prices using STL-LSTM method. 2019 6th International Conference on Systems and Informatics (ICSAI),

Li, F., Zhou, H., Liu, M., & Ding, L. (2023). A Medium to Long-Term Multi-Influencing Factor Copper Price Prediction Method Based on CNN-LSTM. *IEEE Access*, *11*, 69458-69473.

Li, L. (2022). Cross-border E-commerce intelligent information recommendation system based on deep learning. *Computational Intelligence and Neuroscience*, *2022*.

Li, X., Ma, X., Xiao, F., Xiao, C., Wang, F., & Zhang, S. (2022). Time-series production forecasting method based on the integration of Bidirectional Gated Recurrent Unit (Bi-GRU) network and Sparrow Search Algorithm (SSA). *Journal of Petroleum Science and Engineering*, *208*, 109309.

Li, X., Zhang, H., & Zheng, W. (2023). Design of Pricing Decision Algorithm for Cross-Border E-business Import Supply Chain Based on Deep Learning. International Conference on Computational Finance and Business Analytics,

Li, Y. (2022). A cloud computing-based intelligent forecasting method for cross-border E-commerce logistics costs. *Advances in Mathematical Physics*, *2022*, 1-10.

Li, Y., & Pan, Y. (2022). A novel ensemble deep learning model for stock prediction based on stock prices and news. *International Journal of Data Science and Analytics*, *13*(2), 139-149.

Liu, M. (2022). Analysis of cross-border e-commerce commodities in internet of things based on semantic traceability algorithm. *Mathematical Problems in Engineering*, *2022*.

Liu, X., Zhou, R., Qi, D., & Xiong, Y. (2022). A novel methodology for credit spread prediction: depth-gated recurrent neural network with self-attention mechanism. *Mathematical Problems in Engineering*, *2022*.

Luo, S., Ni, Z., Zhu, X., Xia, P., & Wu, H. (2023). A novel methanol futures price prediction method based on multicycle CNN-GRU and attention mechanism. *Arabian Journal for Science and Engineering*, *48*(2), 1487-1501.

Ma, T., Xiang, G., Shi, Y., & Liu, Y. (2022). Horizontal in situ stresses prediction using a CNN-BiLSTM-attention hybrid neural network. *Geomechanics and Geophysics for Geo-Energy and Geo-Resources*, *8*(5), 152.

Malla, J., Lavanya, C., Jayashree, J., & Vijayashree, J. (2022). Bidirectional Gated Recurrent Unit (BiGRU)-Based Bitcoin Price Prediction by News Sentiment Analysis. International Conference on Soft Computing and Signal Processing,

Mehtab, S., & Sen, J. (2020). Stock price prediction using CNN and LSTM-based deep learning models. 2020 International Conference on Decision Aid Sciences and Application (DASA),

Mouthami, K., Yuvaraj, N., & Pooja, R. (2022). Analysis of SARIMA-BiLSTM-BiGRU in Furniture Time Series Forecasting. International Conference on Hybrid Intelligent Systems,

Peltola, A., Barnat, N., Chernova, E., Cristallo, D., Hoffmeister, O., & MacFeely, S. (2022). The role of international organizations in statistical standards setting and outreach: An overview of the UNCTAD contribution. *Statistical Journal of the IAOS*, *38*(2), 501-509.

Rui, C. (2020). Research on classification of cross-border E-commerce products based on image recognition and deep learning. *IEEE Access*, *9*, 108083-108090.

Selvam, P., & Koilraj, J. A. S. (2022). A deep learning framework for grocery product detection and recognition. *Food Analytical Methods*, *15*(12), 3498-3522.

Suman, S., Kaushik, P., Challapalli, S. S. N., Lohani, B. P., Kushwaha, P., & Gupta, A. D. (2022). Commodity Price Prediction for making informed Decisions while trading using Long Short-Term Memory (LSTM) Algorithm. 2022 5th International Conference on Contemporary Computing and Informatics (IC3I),

Tian, S., Li, L., Li, W., Ran, H., Ning, X., & Tiwari, P. (2024). A survey on few-shot class-incremental learning. *Neural Networks*, *169*, 307-324.

Tian, S., Li, W., Ning, X., Ran, H., Qin, H., & Tiwari, P. (2023). Continuous transfer of neural network representational similarity for incremental learning. *Neurocomputing*, *545*, 126300.

Wang, J., Li, F., An, Y., Zhang, X., & Sun, H. (2024). Towards Robust LiDAR-Camera Fusion in BEV Space via Mutual Deformable Attention and Temporal Aggregation. *IEEE Transactions on Circuits and Systems for Video Technology*.

Wang, M., Zhang, Y., Qin, C., Liu, P., & Zhang, Q. (2022). Option pricing model combining ensemble learning methods and network learning structure.

*Mathematical Problems in Engineering*, 2022.

Wang, P.-Y., Chen, C.-T., Su, J.-W., Wang, T.-Y., & Huang, S.-H. (2021). Deep learning model for house price prediction using heterogeneous data analysis along with joint self-attention mechanism. *IEEE Access*, *9*, 55244-55259.

Xu, X., & Zhou, S. (2023). Cross-border e-commerce supply chain decision making considering out-of-stock aversion risk and waste aversion risk. *IEEE Access*.

Yan, L. (2022). Predictive analysis of user behavior processes in cross-border e-commerce enterprises based on deep learning models. *Security and Communication Networks*, *2022*.

Zhao, Z., Liu, Z., Qian, Q., Li, L., Zhao, Y., & Zhu, Y. (2022). Dual-source procurement strategy of cross-border e-commerce supply chain considering members' risk attitude. *Security and Communication Networks*, *2022*.

Liu, Y., Zhao, R., & Li, Y. (2022). A Preliminary Comparison of Drivers' Overtaking behavior between Partially Automated Vehicles and Conventional Vehicles. In Proceedings of the Human Factors and Ergonomics Society Annual Meeting, 66(1), 913-917.

Liu, Y., Zhao, R., Li, T., & Li, Y. (2022). The Impact of Directional Road Signs Combinations and Language Unfamiliarity on Driving Behavior. In International Conference on Human-Computer Interaction, 195-204.

Luo, Y., Xu, N., Peng, H., Wang, C., Duan, S., Mahmood, K., Wen, W., Ding, C., & Xu, X. (2023). AQ2PNN: Enabling Two-party Privacy-Preserving Deep Neural Network Inference with Adaptive Quantization. In 2023 56th IEEE/ACM International Symposium on Microarchitecture (MICRO) (pp. 628-640). IEEE.

Patel, S., Liu, Y., Zhao, R., Liu, X., & Li, Y. (2022). Inspection of in-vehicle touchscreen infotainment display for different screen locations, menu types, and positions. In International Conference on Human-Computer Interaction, 258-279.

Zhu, Z., Zhao, R., Ni, J., & Zhang, J. (2019). Image and spectrum based deep feature analysis for particle matter estimation with weather information. In 2019 IEEE International Conference on Image Processing (ICIP), 3427-3431.

Li, T., Zhao, R., Liu, Y., Li, Y., & Li, G. (2021). Evaluate the effect of age and driving experience on driving performance with automated vehicles. In International Conference on Applied Human Factors and Ergonomics, 155-161.

Song, Y., Fellegara, R., Iuricich, F., & De Floriani, L. (2021). Efficient topology-aware simplification of large triangulated terrains. In Proceedings of the 29th International Conference on Advances in Geographic Information Systems, 576-587.

Jiang, X., Yu, J., Qin, Z., Zhuang, Y., Zhang, X., Hu, Y., & Wu, Q. (2020). Dualvd: An adaptive dual encoding model for deep visual understanding in visual dialogue. In Proceedings of the AAAI Conference on Artificial Intelligence, 34(07), 11125-11132.

An, Z., Wang, X., T. Johnson, T., Sprinkle, J., & Ma, M. (2023). Runtime monitoring of accidents in driving recordings with multi-type logic in empirical models. In International Conference on Runtime Verification, 376-388.

Richardson, A., Wang, X., Dubey, A., & Sprinkle, J. (2024). Reinforcement Learning with Communication Latency with Application to Stop-and-Go Wave Dissipation. In 2024 IEEE Intelligent Vehicles Symposium (IV), 1187-1193.

Wang, X., Onwumelu, S., & Sprinkle, J. (2024). Using Automated Vehicle Data as a Fitness Tracker for Sustainability. In 2024 Forum for Innovative Sustainable Transportation Systems (FISTS), 1-6.

Jiang, L., Yu, C., Wu, Z., Wang, Y., & others. (2024). Advanced AI Framework for Enhanced Detection and Assessment of Abdominal Trauma: Integrating 3D Segmentation with 2D CNN and RNN Models. arXiv preprint arXiv:2407.16165.

De, A., Mohammad, H., Wang, Y., Kubendran, R., Das, A. K., & Anantram, M. P. (2023). Performance analysis of DNA crossbar arrays for high-density memory storage applications. Scientific Reports, 13(1), 6650.

Li, T., Zhao, R., Liu, Y., Liu, X., & Li, Y. (2022). Effect of Age on Driving Behavior and a Neurophysiological Interpretation. In International Conference on Human-Computer Interaction, 184-194.

Wang, C., Sui, M., Sun, D., Zhang, Z., & Zhou, Y. (2024). Theoretical Analysis of Meta Reinforcement Learning: Generalization Bounds and Convergence Guarantees. arXiv preprint arXiv:2405.13290.

Wang, S., Jiang, R., Wang, Z., & Zhou, Y. (2024). Deep Learning-based Anomaly Detection and Log Analysis for Computer Networks. Journal of Information and Computing, 2(2), 34-63.

Jin, C., Peng, H., Zhao, S., Wang, Z., Xu, W., Han, L., Zhao, J., Zhong, K., Rajasekaran, S., & Metaxas, D. N. (2024). APEER: Automatic Prompt Engineering Enhances Large Language Model Reranking. arXiv preprint arXiv:2406.14449.

Liu, Y., Zhao, R., Li, T., & Li, Y. (2021). An investigation of the impact of autonomous driving on driving behavior in traffic jam. In IIE Annual Conference. Proceedings, 986-991.

Xu, Z., Deng, D., Dong, Y., & Shimada, K. (2022). DPMPC-Planner: A real-time UAV

trajectory planning framework for complex static environments with dynamic obstacles. In 2022 International Conference on Robotics and Automation (ICRA), 250-256.

Chen, P., Zhang, Z., Dong, Y., Zhou, L., & Wang, H. (2024). Enhancing Visual Question Answering through Ranking-Based Hybrid Training and Multimodal Fusion. arXiv preprint arXiv:2408.07303.

Zhao, R., Liu, Y., Li, Y., & Tokgoz, B. (2021). An Investigation of Resilience in Manual Driving and Automatic Driving in Freight Transportation System. In IIE Annual Conference. Proceedings, 974-979.

Wang, Y., Alangari, M., Hihath, J., Das, A. K., & Anantram, M. P. (2021). A machine learning approach for accurate and real-time DNA sequence identification. BMC Genomics, 22, 1-10.

Zhou, Y., Wang, Z., Zheng, S., Zhou, L., Dai, L., Luo, H., Zhang, Z., & Sui, M. (2024). Optimization of automated garbage recognition model based on ResNet-50 and weakly supervised CNN for sustainable urban development. Alexandria Engineering Journal, 108, 415-427.

Zhuang, Y., Chen, Y., & Zheng, J. (2020). Music genre classification with transformer classifier. In Proceedings of the 2020 4th International Conference on Digital Signal Processing, 155-159.

De, A., Mohammad, H., Wang, Y., Kubendran, R., Das, A. K., & Anantram, M. P. (2022). Modeling and Simulation of DNA Origami based Electronic Read-only Memory. In 2022 IEEE 22nd International Conference on Nanotechnology (NANO), 385-388.

Jin, C., Huang, T., Zhang, Y., Pechenizkiy, M., Liu, S., Liu, S., & Chen, T. (2023). Visual prompting upgrades neural network sparsification: A data-model perspective. arXiv preprint arXiv:2312.01397.

Peng, X., Xu, Q., Feng, Z., Zhao, H., Tan, L., Zhou, Y., Zhang, Z., Gong, C., & Zheng, Y. (2024). Automatic News Generation and Fact-Checking System Based on Language Processing. arXiv preprint arXiv:2405.10492.

Peng, H., Ran, R., Luo, Y., Zhao, J., Huang, S., Thorat, K., Geng, T., Wang, C., Xu, X., Wen, W., & others. LinGCN: Structural Linearized Graph Convolutional Network for Homomorphically Encrypted Inference. In Thirty-seventh Conference on Neural Information Processing Systems.

Jin, C., Che, T., Peng, H., Li, Y., & Pavone, M. (2024). Learning from teaching regularization: Generalizable correlations should be easy to imitate. arXiv preprint arXiv:2402.02769.

Song, Y., Fellegara, R., Iuricich, F., & De Floriani, L. (2024). Parallel Topology-aware Mesh Simplification on Terrain Trees. ACM Transactions on Spatial Algorithms and Systems, 10(2), 1-39.

Deng, Q., Fan, Z., Li, Z., Pan, X., Kang, Q., & Zhou, M. (2024). Solving the Food-Energy-Water Nexus Problem via Intelligent Optimization Algorithms. arXiv preprint arXiv:2404.06769.

Wang, Y., Khandelwal, V., Das, A. K., & Anantram, M. P. (2022). Classification of DNA Sequences: Performance Evaluation of Multiple Machine Learning Methods. In 2022 IEEE 22nd International Conference on Nanotechnology (NANO), 333-336.

Fellegara, R., Iuricich, F., Song, Y., & Floriani, L. D. (2023). Terrain trees: A framework for representing, analyzing and visualizing triangulated terrains. GeoInformatica, 27(3), 525-564.

Zhou, T., Zhao, J., Luo, Y., Xie, X., Wen, W., Ding, C., & Xu, X. (2024). AdaPI: Facilitating DNN Model Adaptivity for Efficient Private Inference in Edge Computing. arXiv preprint arXiv:2407.05633.

Wang, Y., Demir, B., Mohammad, H., Oren, E. E., & Anantram, M. P. (2023). Computational study of the role of counterions and solvent dielectric in determining the conductance of B-DNA. Physical Review E, 107(4), 044404.

Peng, H., Xie, X., Shivdikar, K., Hasan, M. A., Zhao, J., Huang, S., Khan, O., Kaeli, D., & Ding, C. (2024). MaxK-GNN: Extremely Fast GPU Kernel Design for Accelerating Graph Neural Networks Training. In Proceedings of the 29th ACM International Conference on Architectural Support for Programming Languages and Operating Systems, Volume 2 (pp. 683-698). Association for Computing Machinery, New York, NY, USA.

Wan, Q., Zhang, Z., Jiang, L., Wang, Z., & Zhou, Y. (2024). Image anomaly detection and prediction scheme based on SSA optimized ResNet50-BiGRU model. arXiv preprint arXiv:2406.13987.

Peng, H., Huang, S., Zhou, T., Luo, Y., Wang, C., Wang, Z., Zhao, J., Xie, X., Li, A., Geng, T., & others. (2023). AutoReP: Automatic ReLU Replacement for Fast Private Network Inference. In 2023 IEEE/CVF International Conference on Computer Vision (ICCV) (pp. 5155-5165). IEEE.

Dong, Y. (2024). The Design of Autonomous UAV Prototypes for Inspecting Tunnel Construction Environment. arXiv preprint arXiv:2408.07286.

Zhao, R., Liu, Y., Li, T., & Li, Y. (2022). A Preliminary Evaluation of Driver's Workload in Partially Automated Vehicles. In International Conference on Human-Computer Interaction, 448-458.

Xie, X., Peng, H., Hasan, A., Huang, S., Zhao, J., Fang, H., Zhang, W., Geng, T., Khan, O., & Ding, C. (2023). Accel-gcn: High-performance gpu accelerator design for graph convolution networks. In 2023 IEEE/ACM International Conference on Computer Aided Design (ICCAD) (pp. 01-09). IEEE.

Lee, J. W., Wang, H., Jang, K., Hayat, A., Bunting, M., Alanqary, A., Barbour, W., Fu, Z., Gong, X., Gunter, G., & others. (2024). Traffic smoothing via connected & automated vehicles: A modular, hierarchical control design deployed in a 100-cav flow smoothing experiment. IEEE Control Systems Magazine.